\documentclass[conference]{IEEEtran}
\IEEEoverridecommandlockouts
\usepackage{amsmath,amssymb,amsfonts}
\usepackage{algorithmic}
\usepackage{graphicx}
\usepackage{textcomp}
\usepackage{xcolor}
\usepackage[numbers,sort&compress]{natbib}

\graphicspath{ {figures_arxiv/} }      
\DeclareGraphicsExtensions{.pdf} 
\newcommand{\ie}{{\em i.e.}}
\newcommand{\eg}{{\em e.g.}}

\def\BibTeX{{\rm B\kern-.05em{\sc i\kern-.025em b}\kern-.08em
    T\kern-.1667em\lower.7ex\hbox{E}\kern-.125emX}}
\begin{document}

\title{Design Considerations for Efficient Deep Neural Networks on Processing-in-Memory Accelerators}

\author{Tien-Ju Yang, Vivienne Sze\\
Massachusetts Institute of Technology, Cambridge, USA, email: \{tjy,sze\}@mit.edu
}

\maketitle

\begin{abstract}
This paper describes various design considerations for deep neural networks that enable them to operate efficiently and accurately on processing-in-memory accelerators. We highlight important properties of these accelerators and the resulting design considerations using experiments conducted on various state-of-the-art deep neural networks with the large-scale ImageNet dataset.
\end{abstract}

\section{Introduction}
Deep neural networks (DNNs) deliver state-of-the-art accuracy on a wide range of artificial intelligence tasks at the cost of high computational complexity.  Since data movement tends to dominate energy consumption and can limit throughput for memory-bound workloads, \emph{processing in memory (PIM)} has emerged as a promising way for processing DNNs. When designing efficient DNNs, researchers typically focus on increasing accuracy, and reducing number of weights and multiply-accumulate (MAC) operations; however, it is also critical to account for hardware specific properties when designing DNNs to achieve the desired performance (\eg, accuracy, energy consumption, latency)~\cite{sysml2018-yang}. Unfortunately, the design of efficient DNNs specifically for PIM accelerators has not been widely explored. 

\begin{figure}[!t]
\centering
\includegraphics[width=0.33\textwidth]{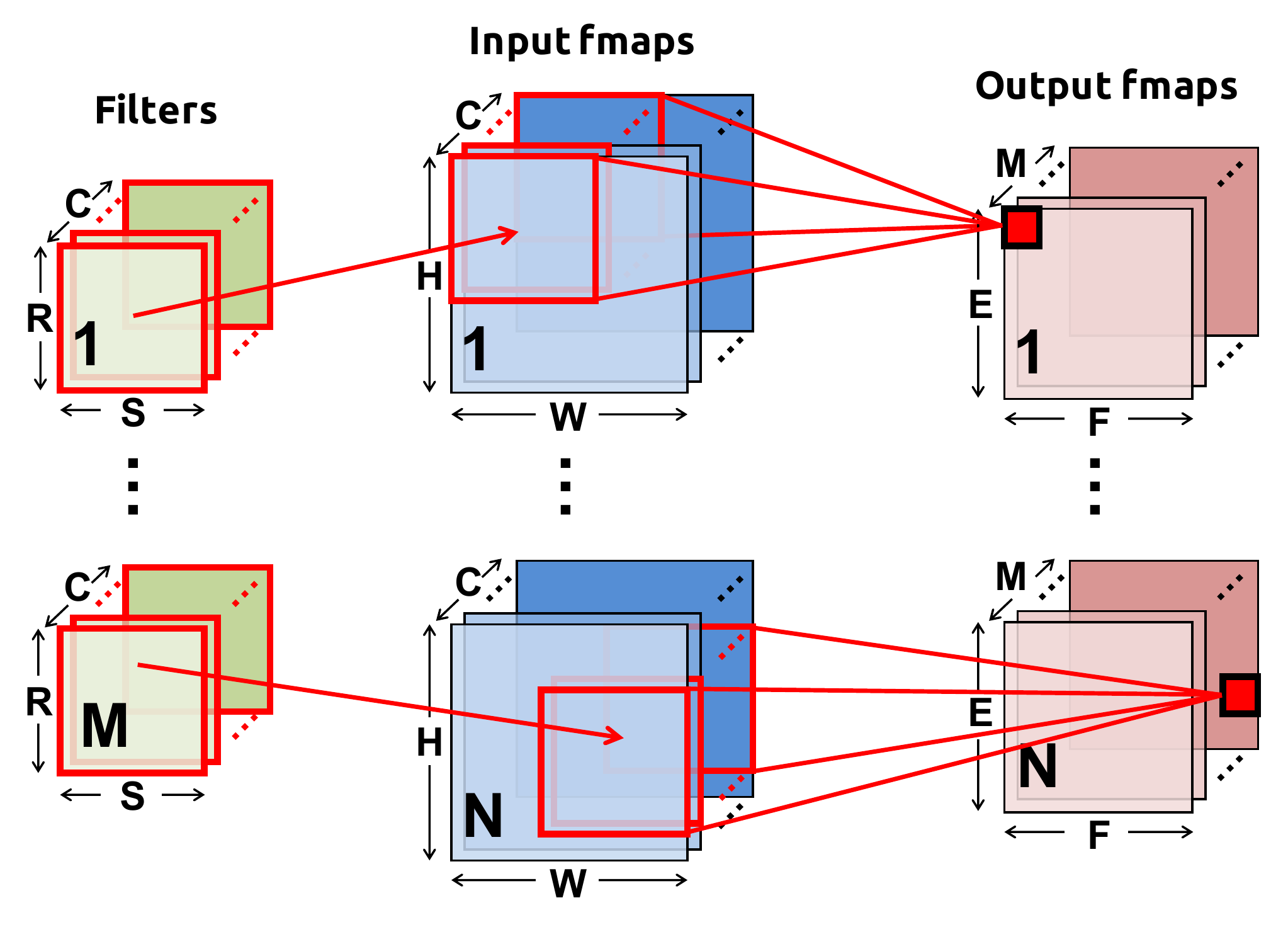}
\caption{An illustration depicting the operations of a convolutional layer and highlighting the terminology adopted in this paper. \textit{fmaps} are feature maps, which contain activations. A fully-connected layer is a special case of a convolutional layer, where $R$ and $S$ are equal to $H$ and $W$.}
\label{fig:cnn_illustraion}
\end{figure}

In this paper, we highlight the key differences between PIM and digital accelerators and summarize how these differences need to be accounted for when designing DNNs for PIM accelerators. The key design considerations include (1) resilience to circuit and device non-idealities, which affect accuracy; (2) data movement of feature map activations, which affects energy consumption and latency; and (3) utilization of the memory array, which affects energy consumption and latency. While there are existing efforts to explore the use of PIM accelerators for running DNNs, most of them target small-scale DNNs -- typically 2 or 3 layer multi-layer perceptrons --  for the simple problem of digit classification on the MNIST dataset~\cite{yu2018neuro}. However, due to the popularity of DNNs, there has been a multitude of DNN architectures proposed in recent years --- with different filter shapes (\ie, $R$, $S$, $C$ and $M$ in Fig.~\ref{fig:cnn_illustraion}) and different number of layers ---  for challenging tasks on larger-scale datasets. These better reveal the challenges for achieving high prediction accuracy and hardware efficiency in practice. In this work, we examine the use of PIM accelerators on 18 recently published DNNs (since 2012) for image classification on the ImageNet dataset to highlight the importance of the various design considerations.

\section{Prediction Accuracy}
To optimize prediction accuracy, we need to consider noise resilience and tolerance to low precision computation. Some works have explored how non-idealities influence accuracy, but they tend to focus more on digital accelerators~\cite{dac2018-reagen} rather than PIM accelerators. We perform our experiments for PIM accelerators assuming the weights cannot be retrained because there are many use cases where the dataset is proprietary. 

\subsection{Noise Resilience}
\label{subsec:noise_resilience}
Compared to digital accelerators, PIM accelerators are more sensitive to non-idealities of the storage devices (\eg, RRAM, PCM, STT-RAM, SRAM bit cell) and the peripheral circuits~\cite{yu2018neuro}. The non-idealities cause the weights and activations to deviate from their intended values and the computation to be noisy, which can significantly impact accuracy. Therefore, the accuracy under the non-idealities should be considered, not just the ``ideal'' accuracy (\ie, accuracy without non-idealities). We model the non-idealities as additive zero-mean Gaussian noise and inject the noise into the output activations of each layer, which conceptually accounts for the noise in the input activations, weights, and computation. To understand whether the type of noise influences the sensitivity of DNNs, we test two types of noise: fixed noise (\ie, the standard deviation of the noise is independent of the magnitude of the activations) and rescaled noise (\ie, the standard deviation of the noise is scaled based on the maximum magnitude of the activations). 

\subsubsection{Fixed Noise}

\begin{figure}[!t]
\centering
\includegraphics[width=0.33\textwidth]{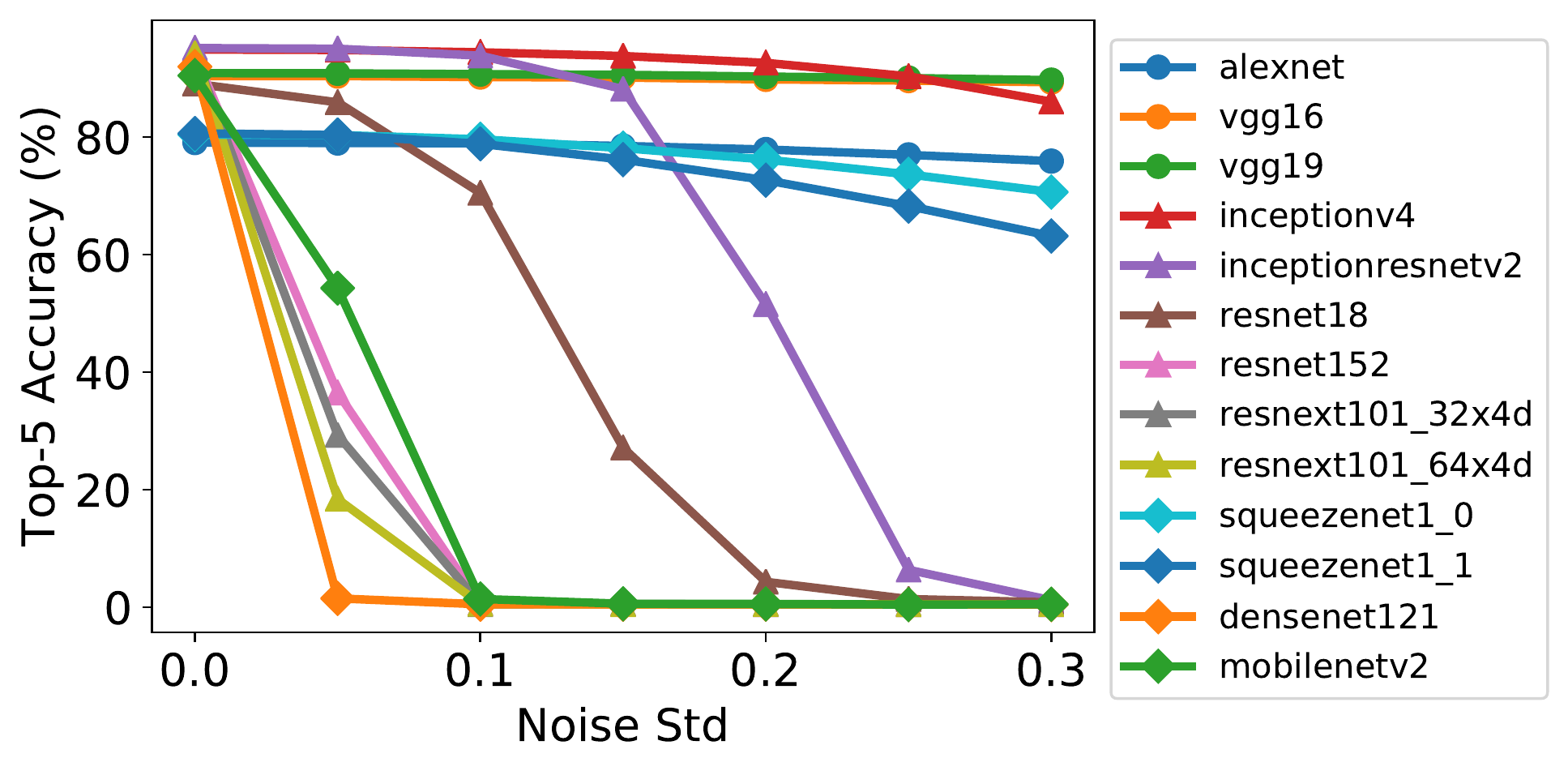}
\caption{The impact of fixed noise in output activations on the accuracy of representative DNNs. The rank of accuracy may change with the standard deviation of the noise.}
\label{fig:classification_ofmap_fixed_all}
\end{figure}

Fig.~\ref{fig:classification_ofmap_fixed_all} shows how the accuracy of different DNNs varies by increasing the standard deviation of the noise. We observe that different DNNs have different sensitivities to noise, but there is no clear relationship between the sensitivity and the ideal accuracy. Specifically, the rank of DNNs in terms of accuracy may change from one noise standard deviation to another. For example, in this experiment, the VGG family becomes the most accurate DNN under strong noise even though it does not achieve the highest ideal accuracy.

\begin{figure}[!t]
\centering
\includegraphics[width=0.33\textwidth]{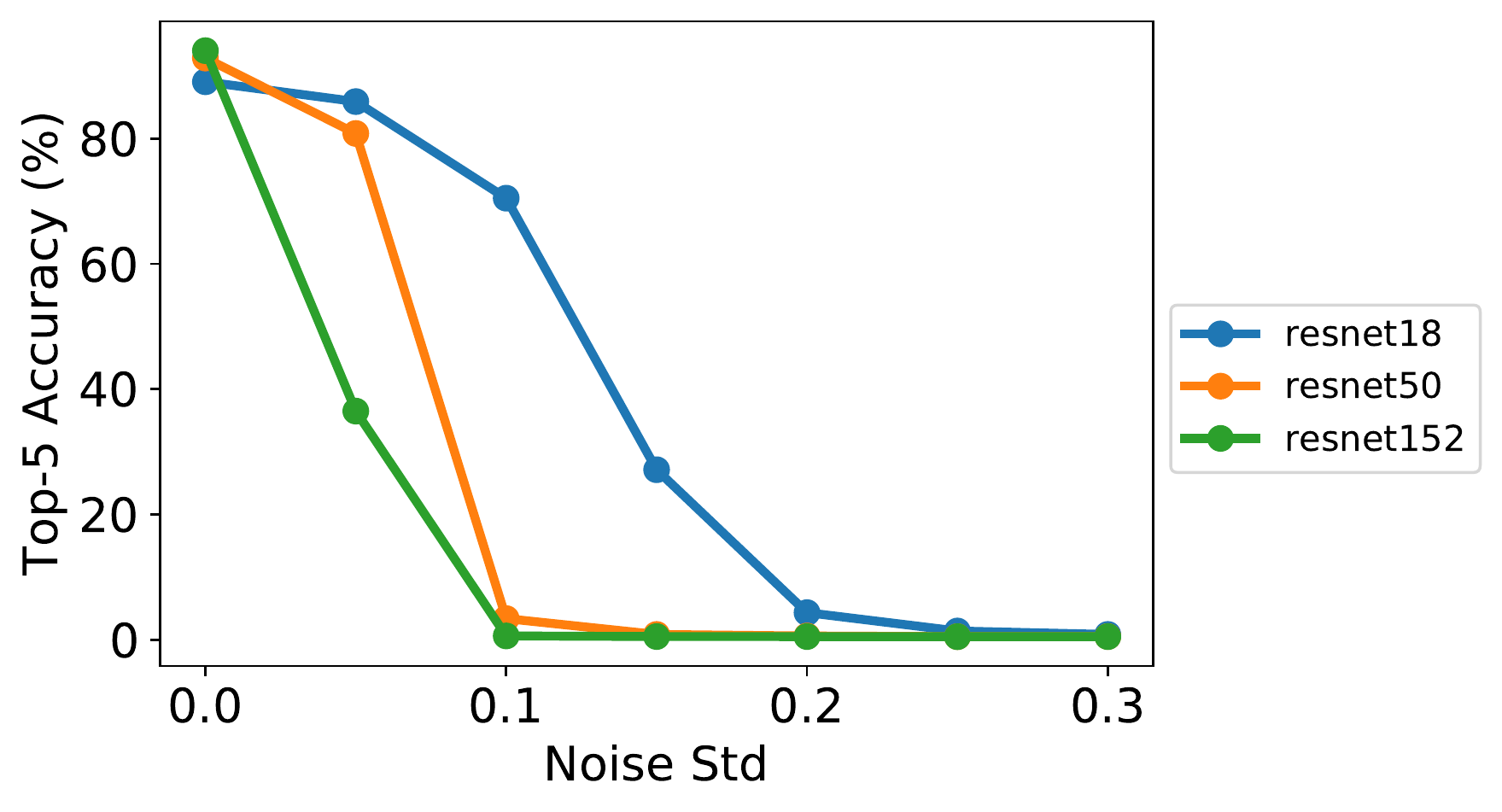}
\caption{The impact of fixed noise in output activations on the accuracy of ResNets with different depths. Reducing the depth of a DNN can make the DNN more robust to noise.}
\label{fig:classification_ofmap_fixed_depth}
\end{figure}

\begin{figure}[!t]
\centering
\includegraphics[width=0.33\textwidth]{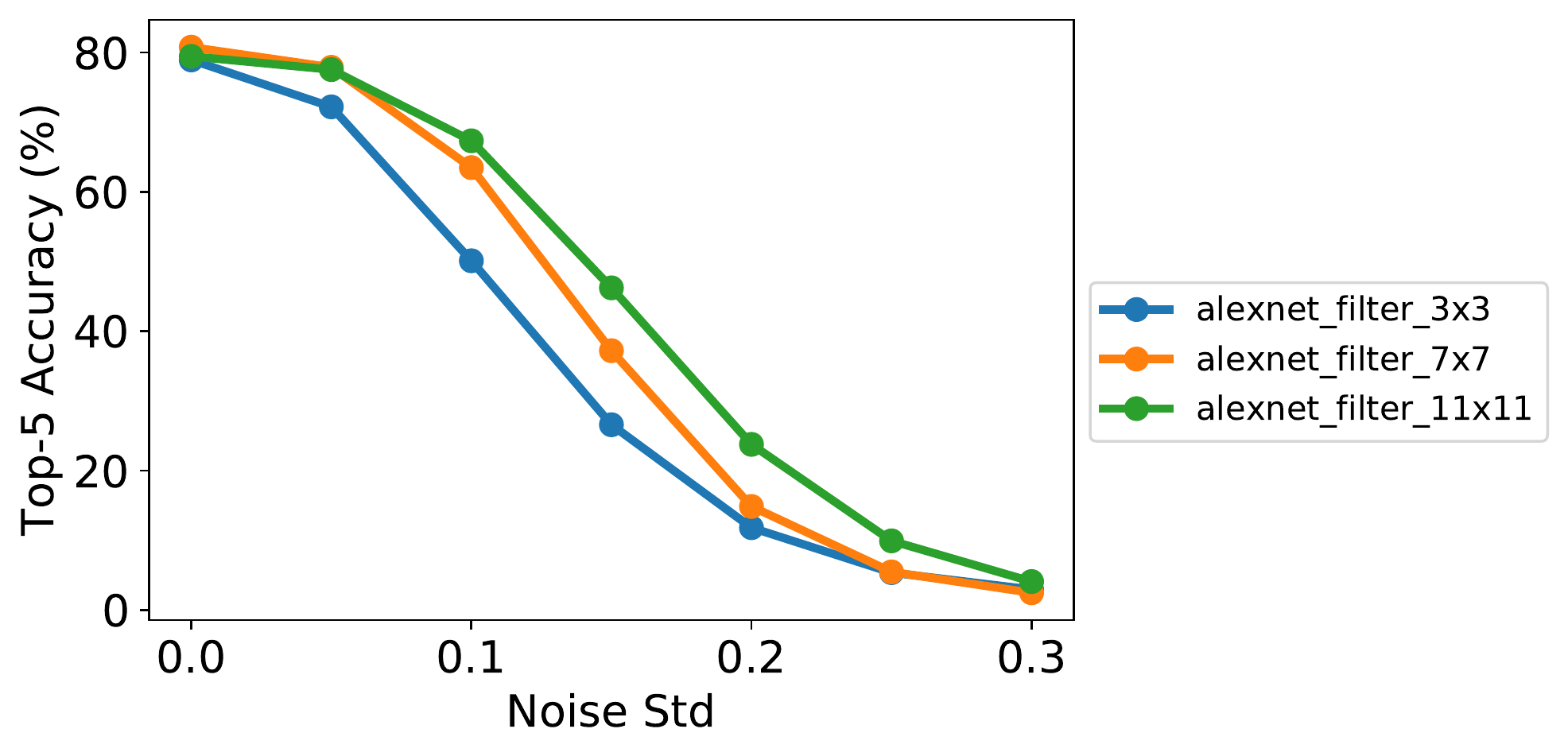}
\caption{The impact of fixed noise in output activations on the accuracy of AlexNet variants with different filter sizes. Increasing the filter size of a DNN can make the DNN more robust to noise.}
\label{fig:classification_ofmap_fixed_kernel}
\end{figure}

Recently, the VGG-like DNNs are less frequently used and replaced by DNNs that are deeper (\ie, more layers) and narrower (\ie, smaller layers). This design approach successfully reduces the number of MACs and weights without lowering accuracy when there are no non-idealities. To evaluate this design approach for PIM accelerators, we analyze the impact of depth and filter size on the accuracy at various noise standard deviations. Fig.~\ref{fig:classification_ofmap_fixed_depth} shows how the accuracy varies with depth by comparing ResNets with 18, 50, 152 layers. In this case, the ideal accuracy increases as the depth increases. However, the accuracy of deeper DNNs decreases faster with increased noise and eventually becomes lower than that of shallower DNNs. Fig.~\ref{fig:classification_ofmap_fixed_kernel} shows how the accuracy varies with filter size by comparing AlexNet variants with filter sizes of $3\times3$, $7\times7$ and $11\times11$. It shows that the accuracy of smaller filters decreases faster with noise. Therefore, the recent trend of making a DNN deeper with smaller layers may not be the most suitable for PIM.

\subsubsection{Rescaled Noise}

\begin{figure}[!t]
\centering
\includegraphics[width=0.33\textwidth]{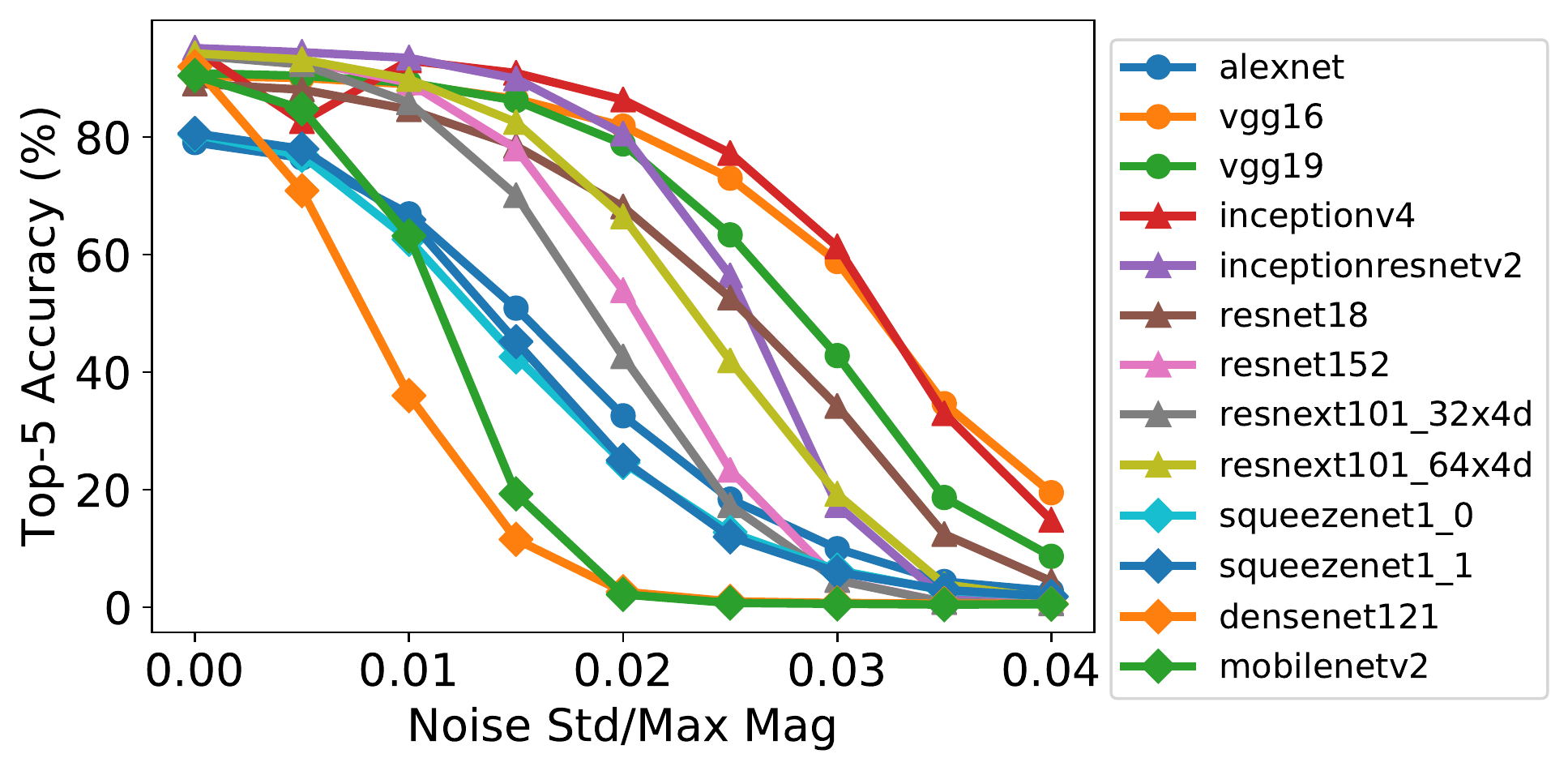}
\caption{The impact of rescaled noise in output activations on the accuracy of representative DNNs. The rank of accuracy may change with the ratio of the standard deviation of the noise to the maximum magnitude of the activations.}
\label{fig:classification_ofmap_relative_max_all}
\end{figure}

\begin{figure}[!t]
\centering
\includegraphics[width=0.33\textwidth]{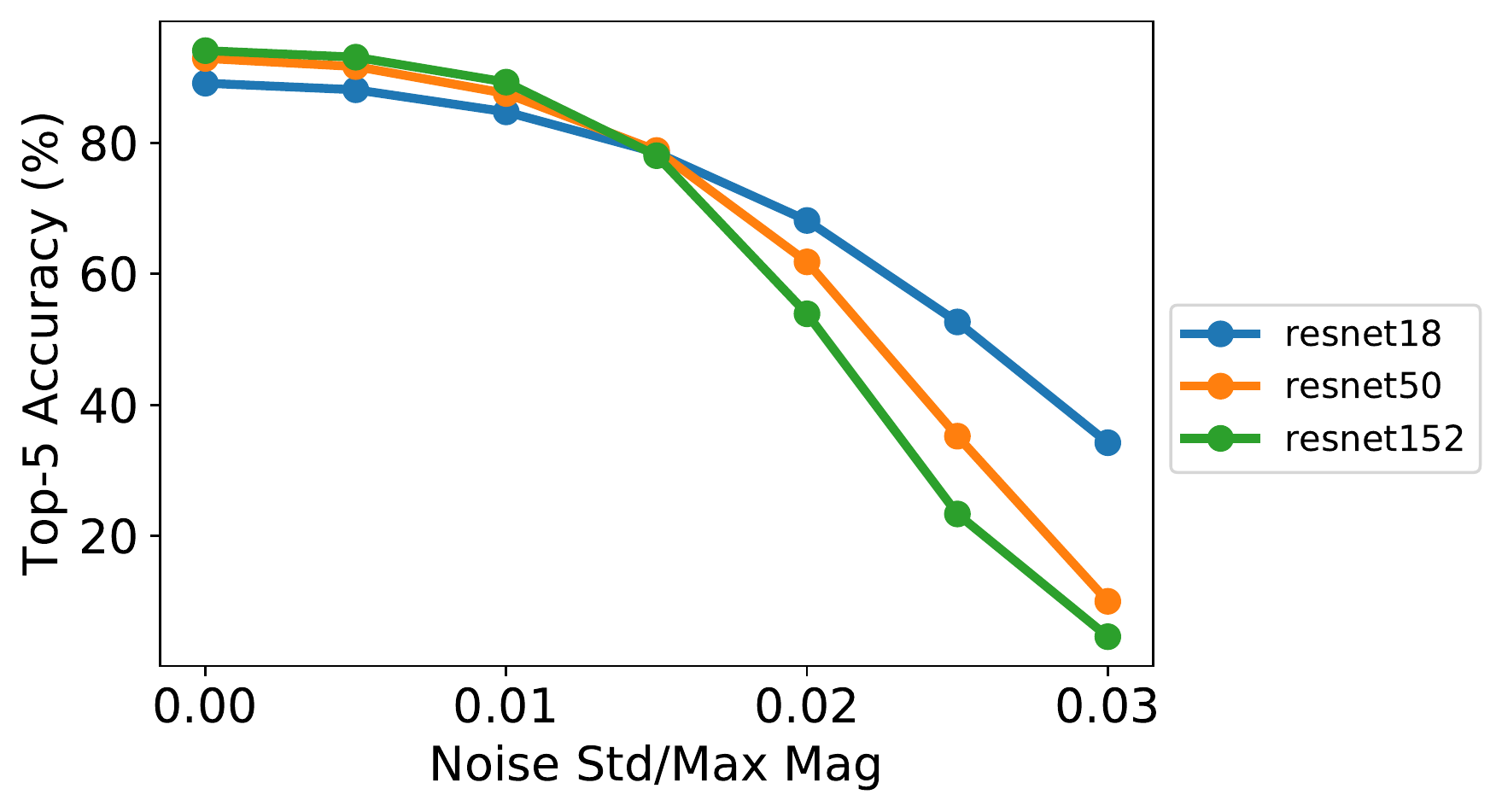}
\caption{The impact of rescaled noise in output activations on the accuracy of ResNets with different depths. Reducing the depth of a DNN can make the DNN more robust to noise.}
\label{fig:classification_ofmap_relative_max_depth}
\end{figure}

\begin{figure}[!t]
\centering
\includegraphics[width=0.33\textwidth]{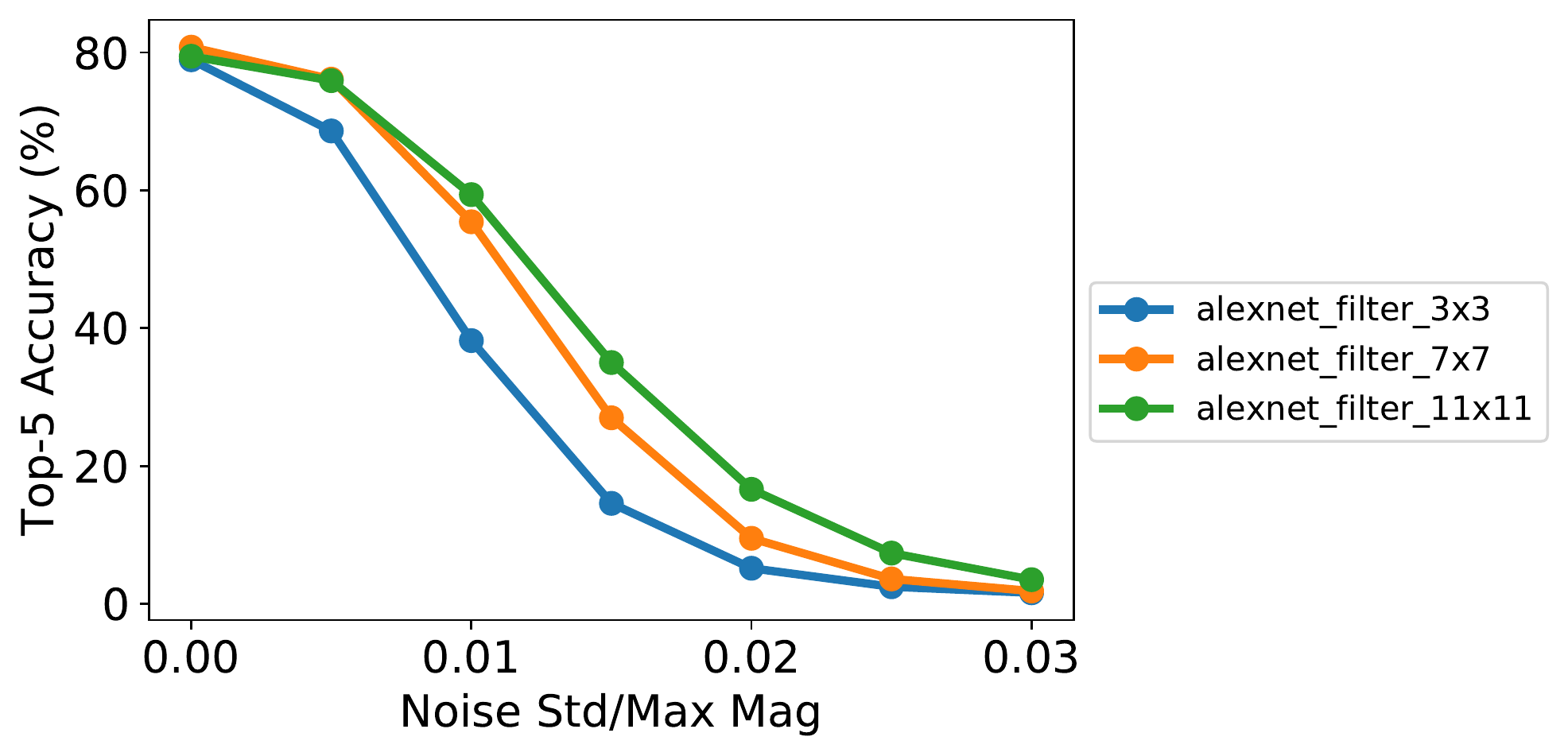}
\caption{The impact of rescaled noise in output activations on the accuracy of AlexNet variants with different filter sizes. Increasing the filter size of a DNN can make the DNN more robust to noise.}
\label{fig:classification_ofmap_relative_max_kernel}
\end{figure}

In this experiment, we rescale the noise relative to the per-layer maximum magnitude of the activations. This can occur if the PIM accelerator has a limited dynamic range for the input activations due to the peripheral circuits~\cite{frontiers2016-RPU}; as a result, the input activations need to be \emph{rescaled} to a predefined range before being streamed into the accelerator. Conceptually, the magnitude of the noise is proportional to the maximum magnitude of the activations.

Fig.~\ref{fig:classification_ofmap_relative_max_all} shows how the accuracy of different DNNs varies by increasing the ratio of the standard deviation of the noise to the maximum magnitude of the activations. The rank in terms of accuracy again changes with noise levels. Although the VGG family is still the most robust to noise, the  change in rank as the noise increases (\ie, sensitivity) is different from that in the fixed-noise experiment (Fig.~\ref{fig:classification_ofmap_fixed_all}). Therefore, the type of noise should be considered when designing DNNs. Fig.~\ref{fig:classification_ofmap_relative_max_depth} and Fig.~\ref{fig:classification_ofmap_relative_max_kernel} repeat the experiments on depth and filter size. We observe that they follow the same trend as the fixed-noise experiments: reducing the depth or increasing the filter size of a DNN can make the DNN more robust to noise.

\subsection{Low Precision Computation}
\label{sec:low_precision}

\begin{figure}[!t]
\centering
\includegraphics[width=0.33\textwidth]{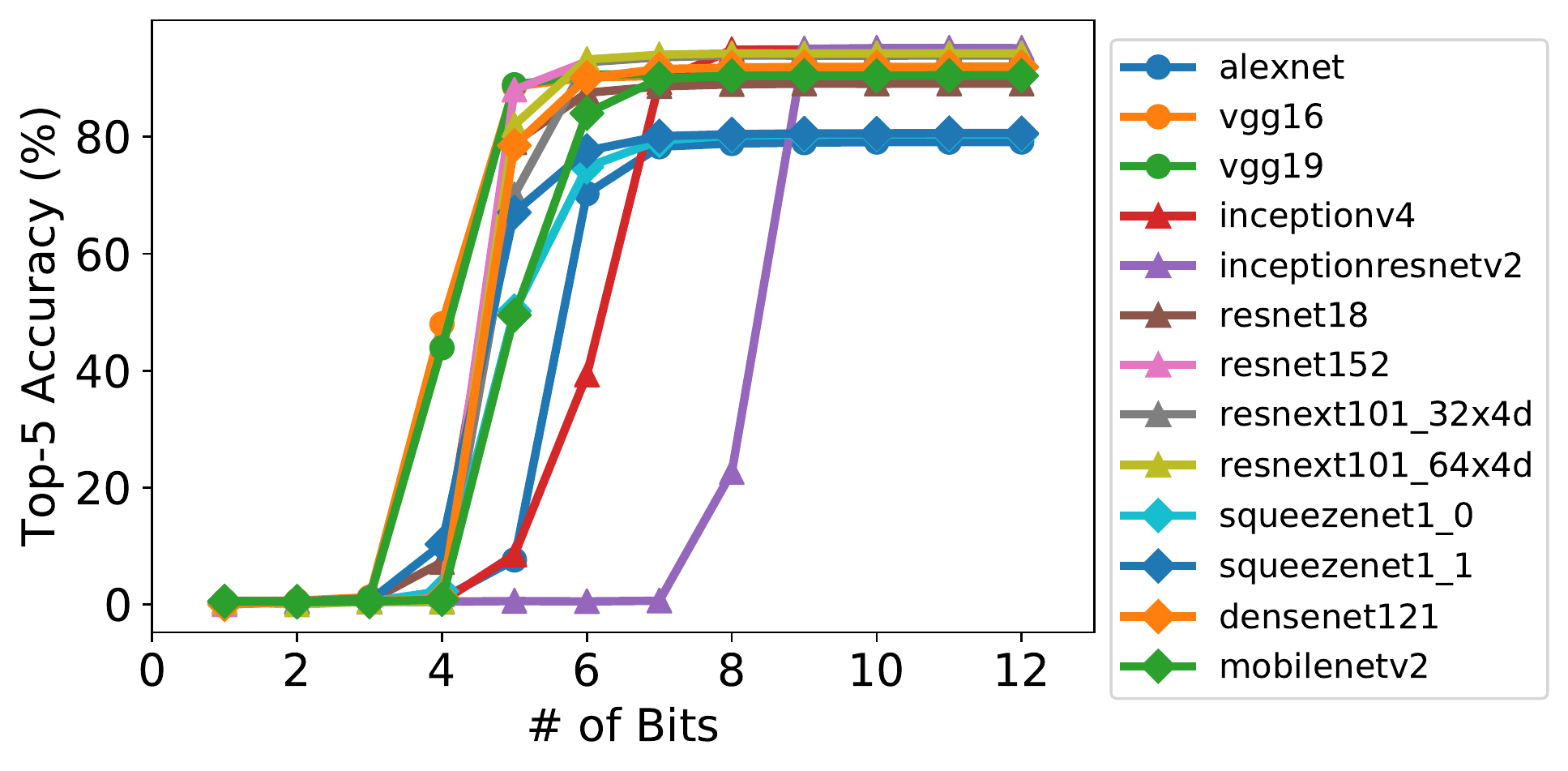}
\caption{The impact of the bit widths of the weights on the accuracy of representative DNNs. The rank in terms of accuracy may change with bit width reduction.}
\label{fig:classification_weight_quantize_all}
\end{figure}

Compared to digital accelerators, PIM accelerators tend to use lower precision for data and computation~\cite{yu2018neuro}.  Fig.~\ref{fig:classification_weight_quantize_all} shows how the accuracy of different DNNs varies with lower weight precision (\ie, lower bit widths). Similarly, we see that the DNNs that start with the highest accuracy at large bit widths may achieve lower accuracy with a reduced bit width. Moreover, the shallower DNNs with larger layers (\eg, VGG) achieve the highest accuracy at 4 bits in this experiment. This trend is similar to what we observed for noise in Sec.~\ref{subsec:noise_resilience}.

\subsection{Summary}
When designing high accuracy DNNs for PIM accelerators, the sensitivity to noise and low precision computation should be taken into account rather than only the ideal accuracy. Moreover, reducing the depth or increasing the filter size of a DNN can help address these challenges. We hypothesize that larger layers with more weights have more redundancy and shallower DNNs have less accumulated errors across the layers of the DNN, which makes them more robust to noise.

\section{Hardware Efficiency}
When designing DNNs to achieve optimal energy efficiency and latency, we need to consider the data movement of activations and their utilization for various memory array sizes.

\subsection{Data Movement of Activations}

\begin{figure}[!t]
\centering
\includegraphics[width=0.21\textwidth]{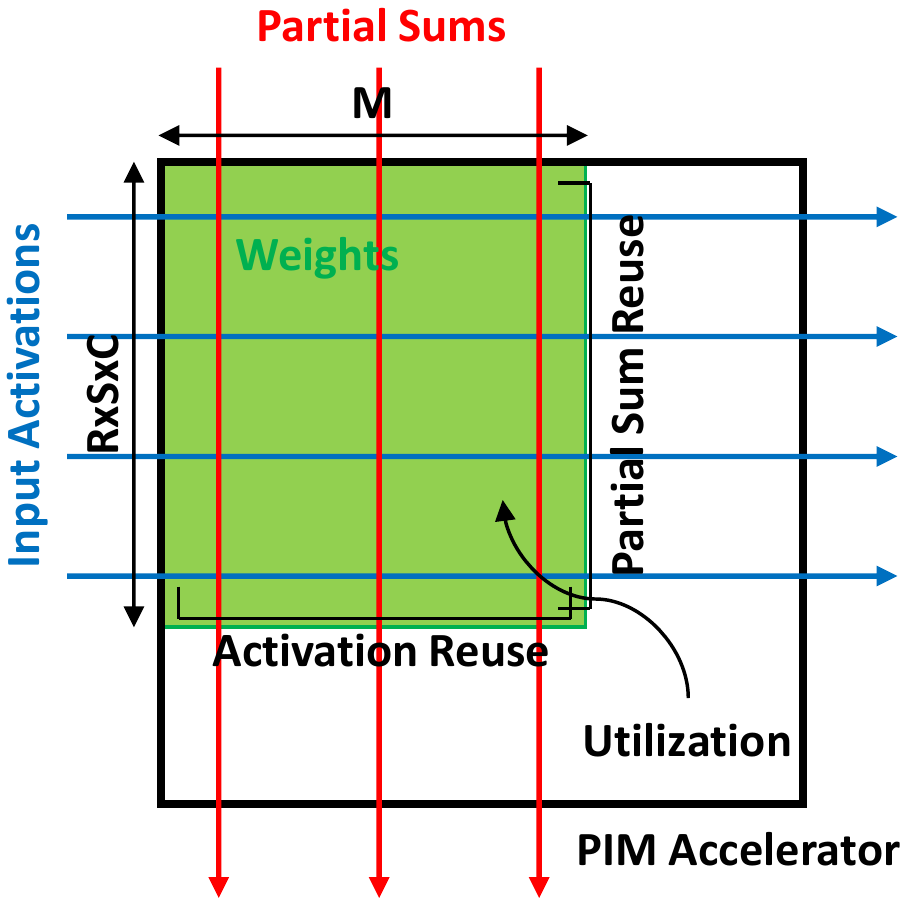}
\caption{The illustration of reuse and utilization for PIM accelerators, where $R$ and $S$ are the height and width of a filter, $C$ is the number of input channels, and $M$ is the number of output channels (refer to Fig.~\ref{fig:cnn_illustraion}). Reuse refers to how many times a value (\eg, activation, partial sum) is used when it moves into the array.}
\label{fig:pim_illustration}
\end{figure}

Fig.~\ref{fig:pim_illustration} shows a simplified illustration of a PIM accelerator. Since PIM brings the computation into the memory where the weights are stored, the data movement of the weights is significantly reduced; this is commonly referred to as a \emph{weight-stationary} dataflow~\cite{isca2016-chen}. However, activations that move in and out of the memory can dominate energy consumption of PIM accelerators due to the costly peripheral circuits. For example, the power of 9-bit SAR-ADCs is roughly 1 W compared to 0.3 W dissipated on a $4096\times4096$ array due to MAC operations~\cite{frontiers2016-RPU}. Therefore, while prior work often focuses only on reducing the number of weights and MACs, for PIM, we also need to consider the number of activations.

\begin{figure}[!t]
\centering
\includegraphics[width=0.33\textwidth]{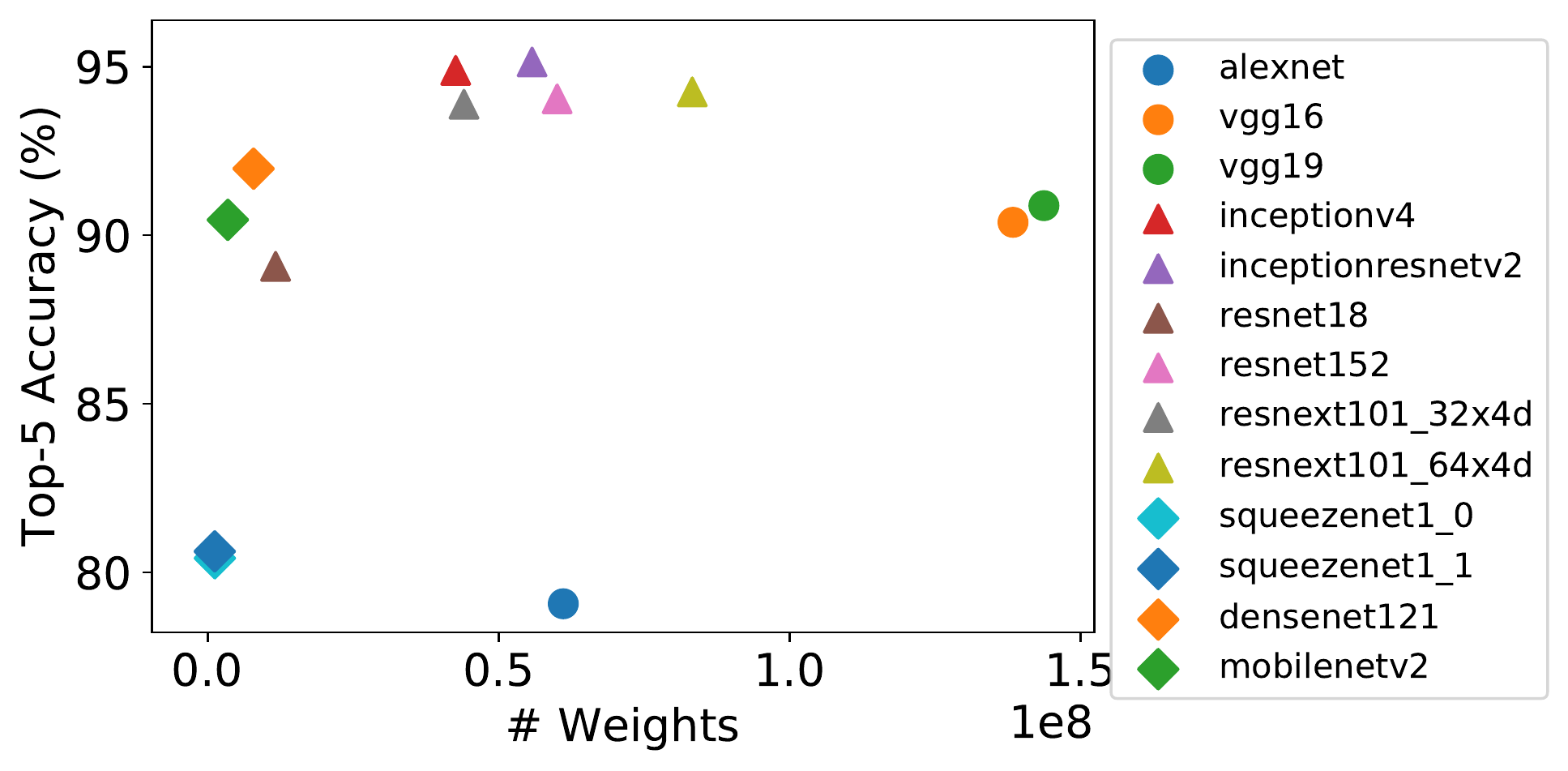}
\caption{The relationship between accuracy and number of weights for representative DNNs. DNNs with fewer weights can have more activations (refer to Fig.~\ref{fig:model_analysis_classification_num_oacts}).}
\label{fig:model_analysis_classification_num_weights}
\end{figure}

\begin{figure}[!t]
\centering
\includegraphics[width=0.33\textwidth]{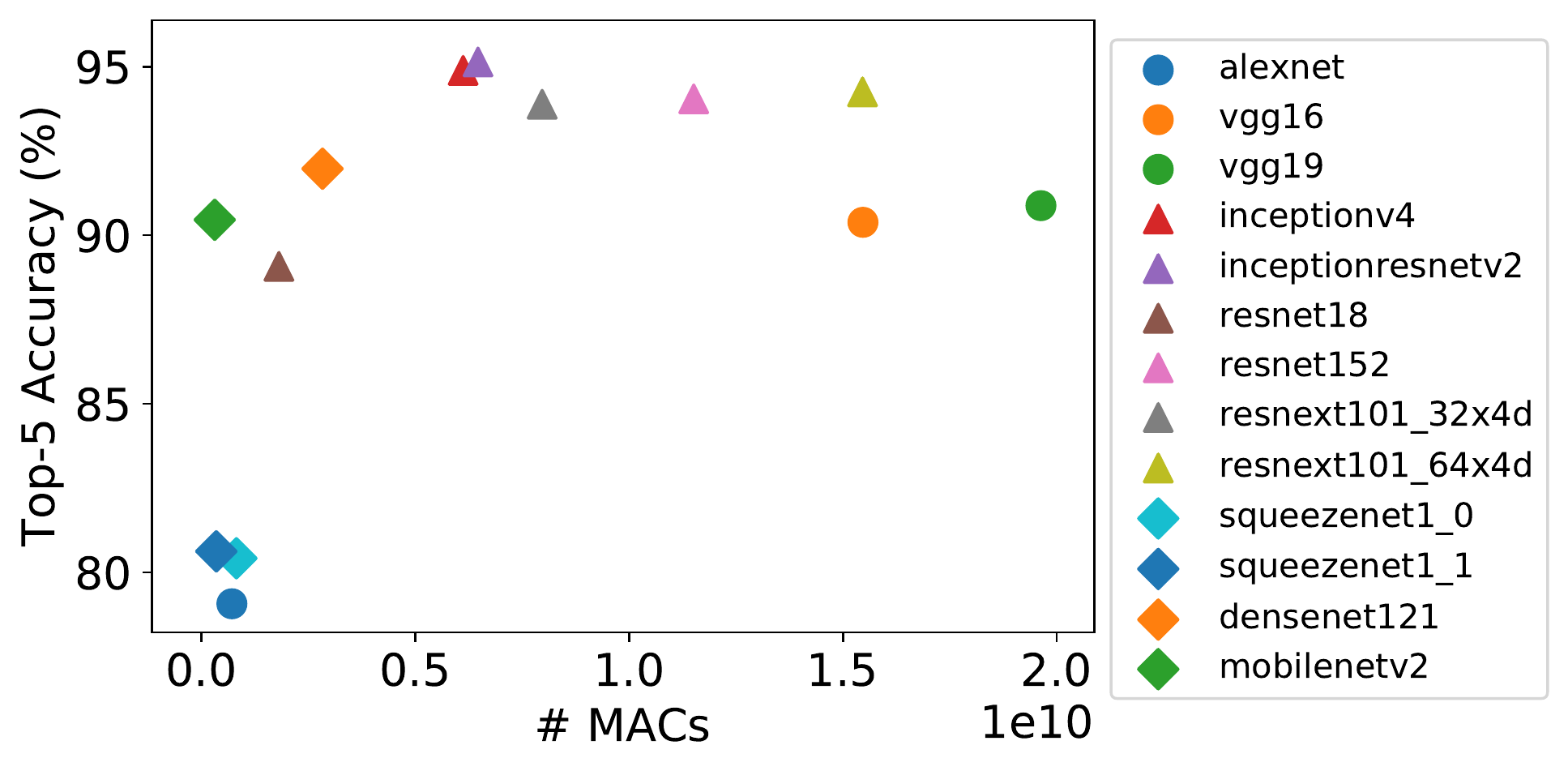}
\caption{The relationship between accuracy and number of MACs for representative DNNs. DNNs with fewer MACs can have more activations (refer to Fig.~\ref{fig:model_analysis_classification_num_oacts}).}
\label{fig:model_analysis_classification_num_macs}
\end{figure}

\begin{figure}[!t]
\centering
\includegraphics[width=0.33\textwidth]{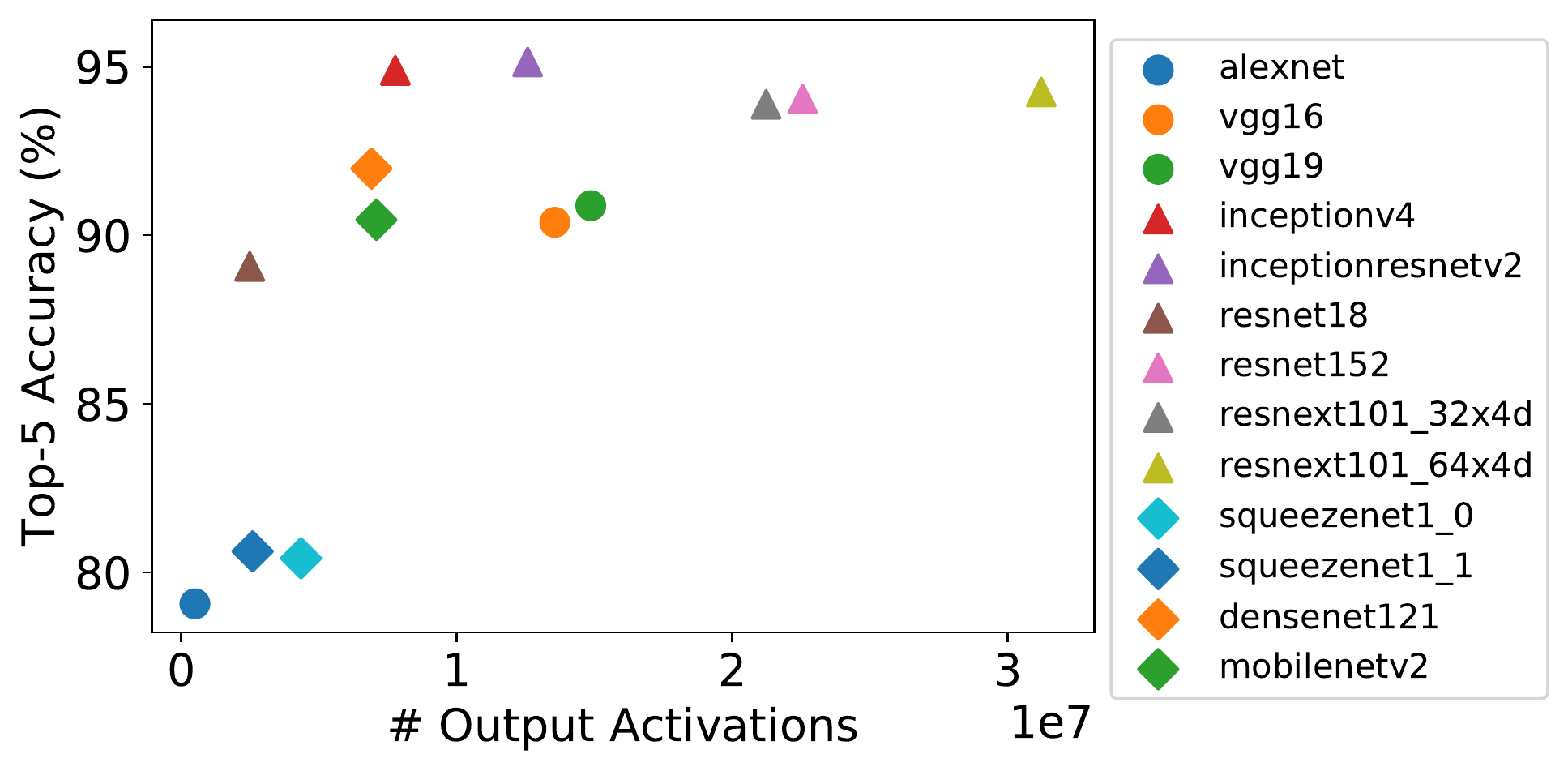}
\caption{The relationship between accuracy and number of output activations for representative DNNs. DNNs with fewer weights and fewer MACs can have more activations (refer to Fig.~\ref{fig:model_analysis_classification_num_weights} and Fig.~\ref{fig:model_analysis_classification_num_macs}).}
\label{fig:model_analysis_classification_num_oacts}
\end{figure}

Fig.~\ref{fig:model_analysis_classification_num_weights}, Fig.~\ref{fig:model_analysis_classification_num_macs}, and Fig.~\ref{fig:model_analysis_classification_num_oacts} show the trade-off between the accuracy versus the number of weights, MACs and activations for different DNNs. When comparing DNNs with similar accuracies, we observe that a decrease in the number of MACs and weights can be accompanied by an increase in the number of activations, which may offset the energy benefit from the reduced number of weights and MACs.

In addition to the number of activations, the reuse (\eg, activation reuse in Fig.~\ref{fig:pim_illustration}) is another critical factor in data movement. Reuse refers to how many times a value is used (\eg, an activation is used by multiple weights) when it moves into the array. When a value is used more times, the number of reads from outside the array is reduced. The array utilization influences the amount of reuse that can be exploited by the hardware, and we will discuss it in the next section.

\subsection{Impact of Array Size on Utilization}

Compared to digital accelerators, PIM accelerators tend to have a larger array size since the energy and area costs of the peripheral circuits are much larger than the memory array; the array size of digital accelerators typically range from $16\times16$ to $128\times128$~\cite{pieee2017-sze} compared with $128\times128$ to $4096\times4096$ for PIM accelerators~\cite{yu2018neuro, frontiers2016-RPU}. The latency of PIM accelerators is determined not only by the number of MACs but also by how many MACs are processed in parallel based on the array utilization; thus low utilization can cause longer latency on PIM accelerators.

\begin{figure}[!t]
\centering
\includegraphics[width=0.33\textwidth]{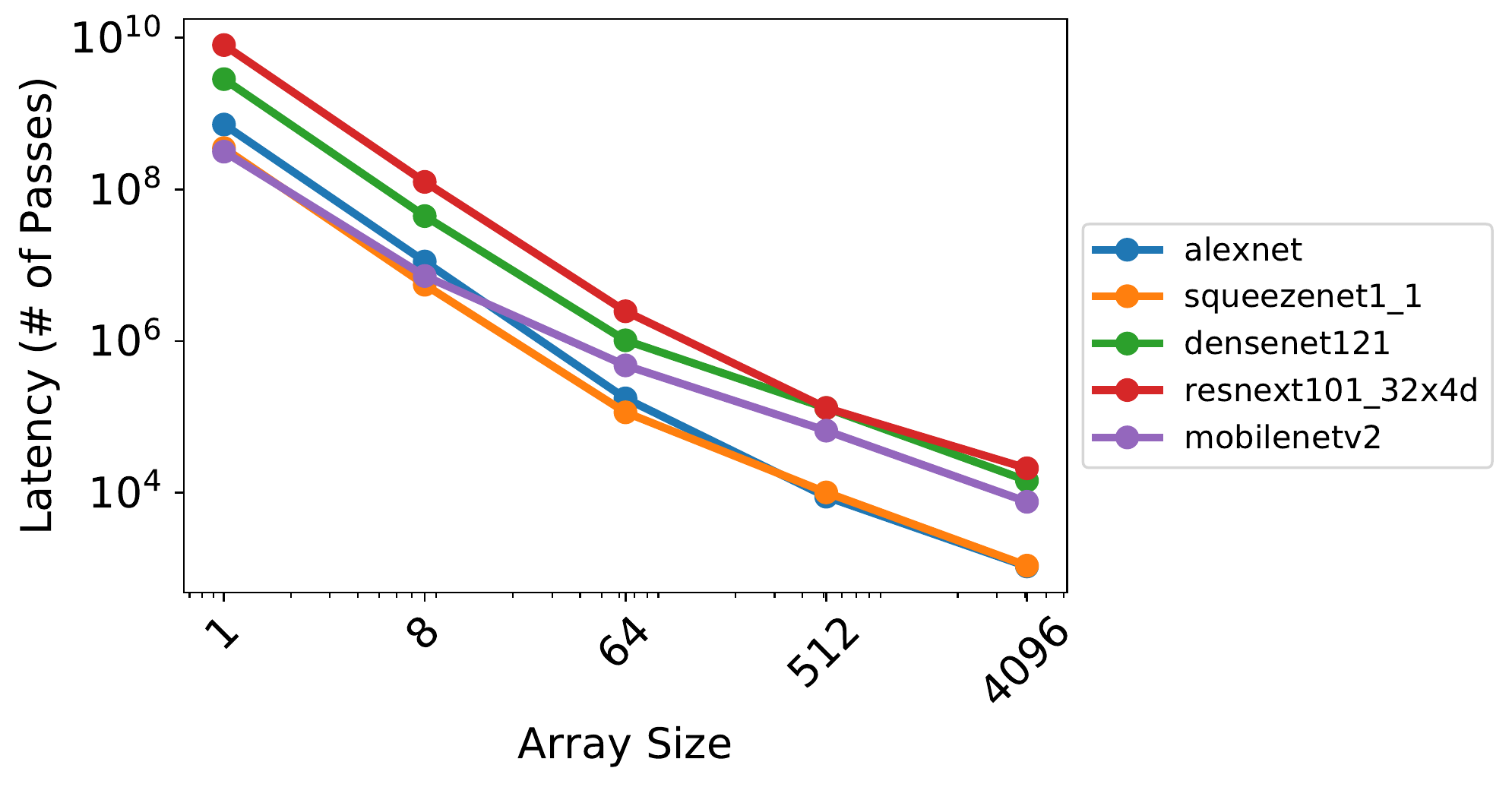}
\caption{The impact of array size on the estimated latency for various DNNs. Shallower DNNs with larger layers (\eg, AlexNet) benefit more from a large array than deeper DNNs with smaller layers.}
\label{fig:model_analysis_num_passes_all}
\end{figure}

Utilization also influences energy consumption. Higher utilization can provide more opportunities to reuse data. Fig.~\ref{fig:model_analysis_num_passes_all} shows that shallower DNNs with larger layers (\eg, AlexNet) may benefit more from a large array. This is because deeper DNNs with smaller layers focus on reducing the number of MACs and weights, which maps better to smaller arrays. Therefore, when designing a DNN, its utilization needs to be considered to minimize latency and energy consumption.

\subsection{Trade-Off between Layer Size and DNN Depth}

\begin{figure}[!t]
\centering
\includegraphics[width=0.33\textwidth]{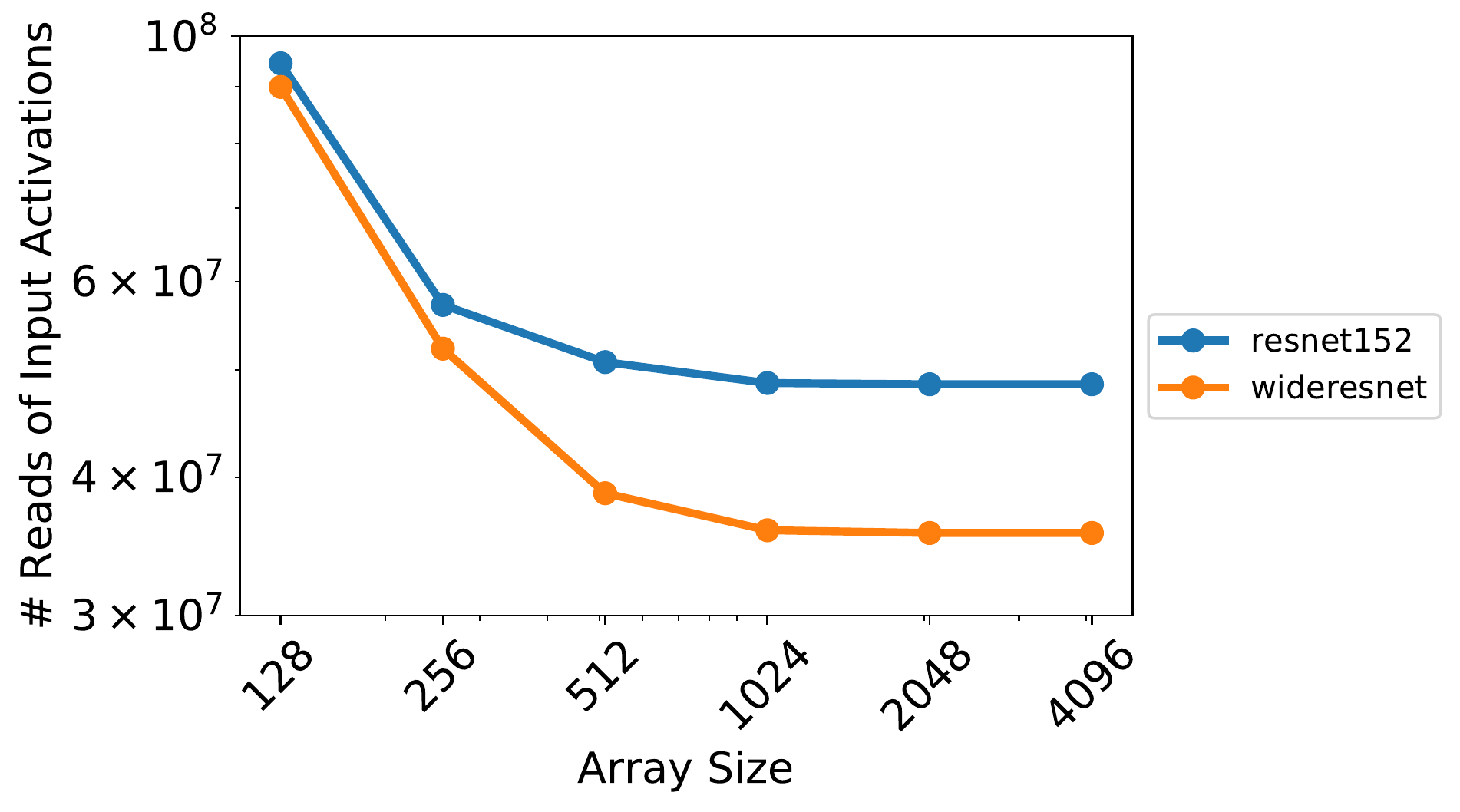}
\caption{The impact of array size on the estimated number of reads of input activations for ResNet152 and Wide ResNet. Decreasing the depth and increasing the layer size can reduce the data movement of activations without lowering the accuracy. It becomes more effective when the array becomes larger.}
\label{fig:model_analysis_num_read_iacts_wideresnet}
\end{figure}

\begin{figure}[!t]
\centering
\includegraphics[width=0.33\textwidth]{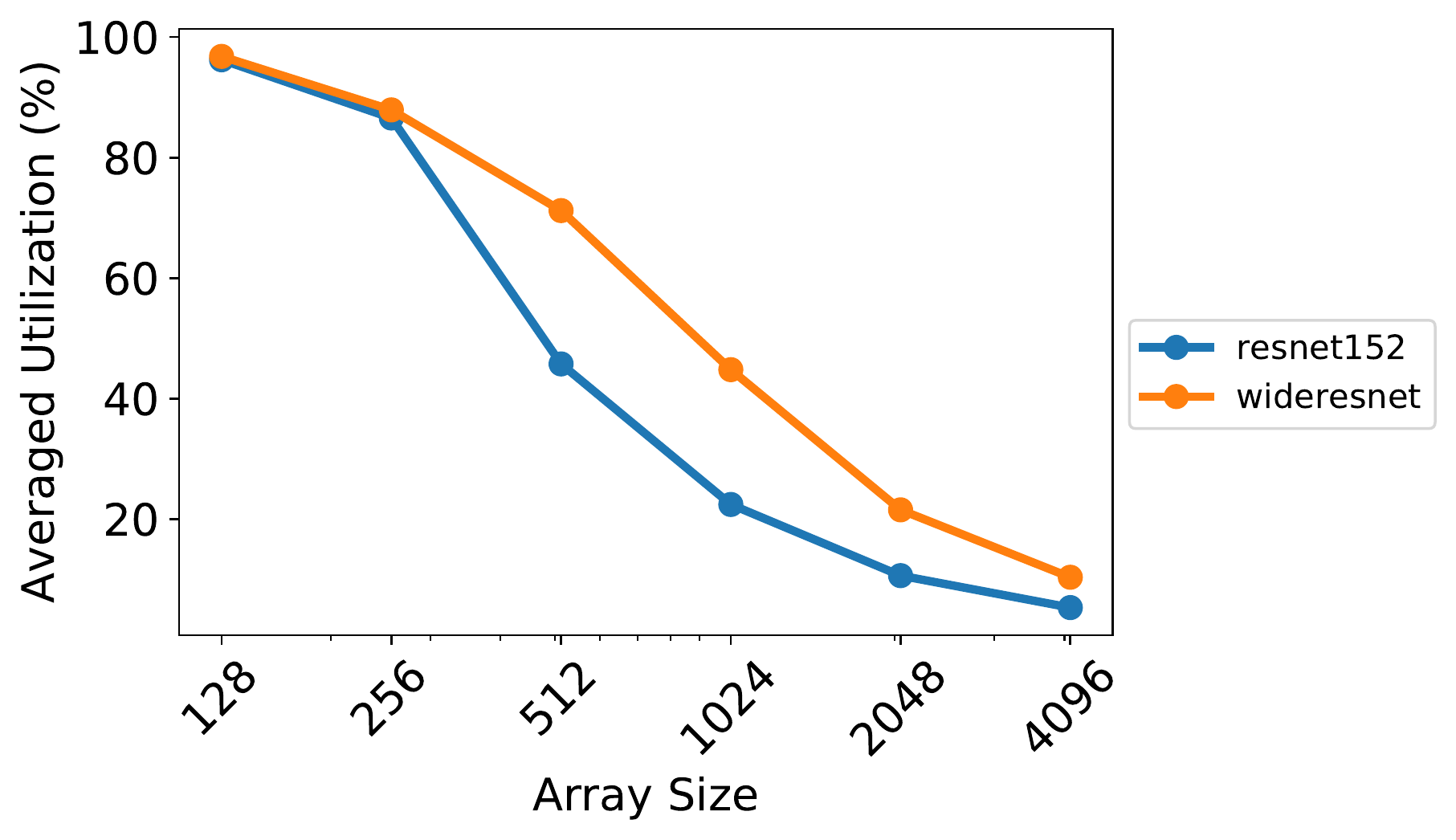}
\caption{The impact of array size on the averaged estimated utilization for ResNet152 and Wide ResNet. Decreasing the depth and increasing the layer size can increase the utilization of a large array without lowering the accuracy.}
\label{fig:model_analysis_utilization_wideresnet}
\end{figure}

\begin{figure}[!t]
\centering
\includegraphics[width=0.33\textwidth]{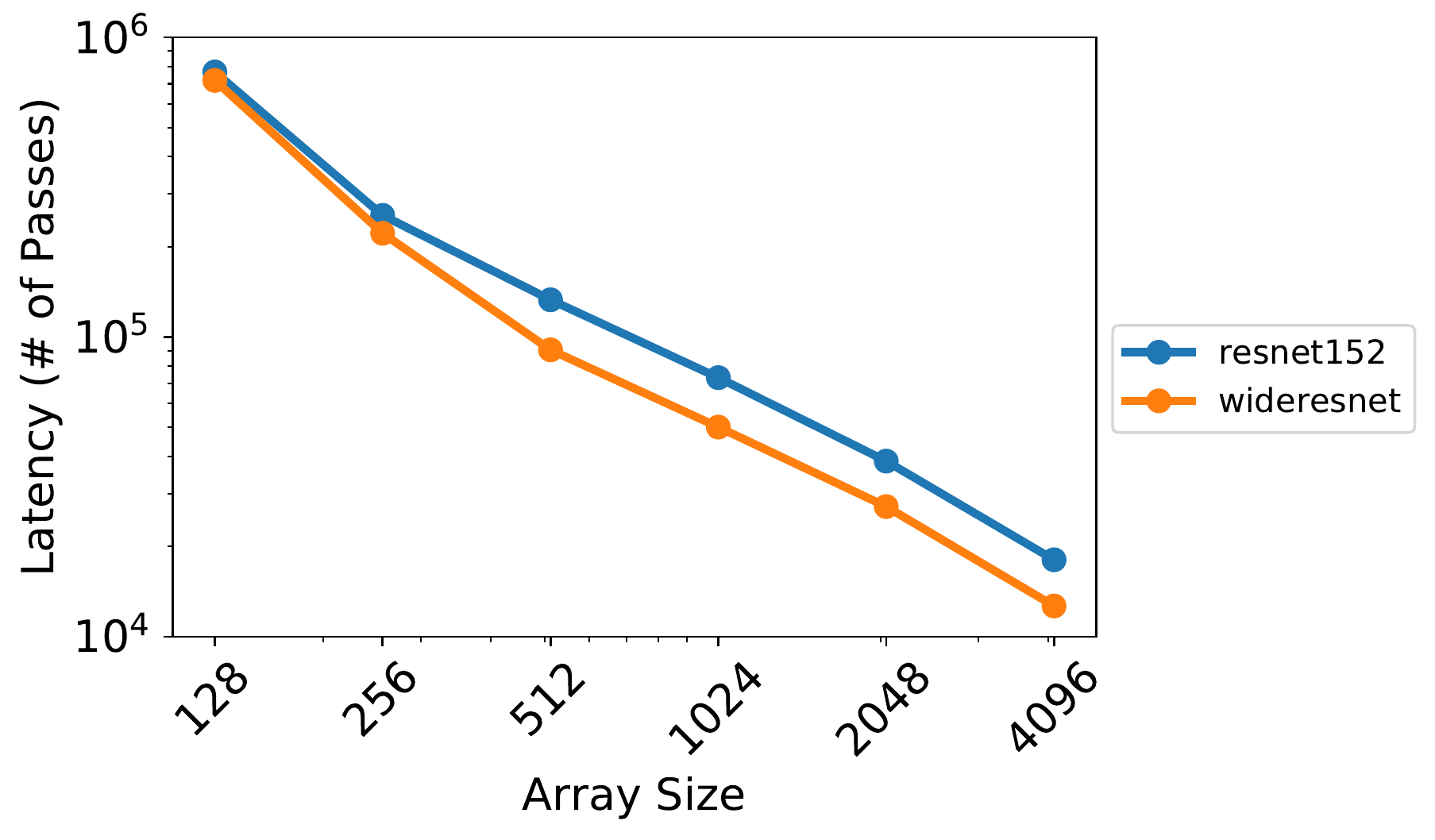}
\caption{The impact of array size on the estimated latency for ResNet152 and Wide ResNet. Decreasing the depth and increasing the layer size can reduce the latency when using a large array without lowering the accuracy.}
\label{fig:model_analysis_num_passes_wideresnet}
\end{figure}

One approach to increasing the hardware efficiency of a DNN for PIM accelerators without lowering the accuracy is to reduce the depth and increase the layer size. For example, Wide ResNet~\cite{zagoruyko2016-wideresnet} achieves a similar accuracy and number of MACs as ResNet152 with fewer but larger layers. The smaller depth reduces the number of activations, and the larger layers enable more activation reuse in PIM, which results in a lower number of activation reads on large arrays as shown in Fig.~\ref{fig:model_analysis_num_read_iacts_wideresnet}. The larger layers also enable higher utilization (Fig.~\ref{fig:model_analysis_utilization_wideresnet}) and lead to shorter latency (Fig.~\ref{fig:model_analysis_num_passes_wideresnet}).

\section{Conclusion}
This paper summarizes important design considerations when designing DNNs for PIM accelerators to achieve higher prediction accuracy and hardware efficiency. These considerations include the sensitivity to non-idealities, the activations, and the utilization in addition to the weights and MACs and the ideal accuracy. Moreover, making DNNs shallower with larger layers may not result in accuracy degradation while producing efficient DNN designs suitable for PIM accelerators, but it remains an open question for the community to explore.

\section*{Acknowledgment}
We thank Tayfun Gokmen for helpful discussions. This work was supported by the MIT-IBM Watson AI Lab, the MIT Quest for Intelligence, and by the NSF E2CDA 1639921.

{\footnotesize
\bibliographystyle{ieeetr}
\bibliography{__references}

\begin{thebibliography}{1}

\bibitem{sysml2018-yang}
Y.-H. Chen, T.-J. Yang, J.~Emer, and V.~Sze, ``{Understanding the Limitations
  of Existing Energy-Efficient Design Approaches for Deep Neural Networks},''
  in {\em SysML Conference}, 2018.

\bibitem{yu2018neuro}
S.~Yu, ``{Neuro-Inspired Computing with Emerging Nonvolatile Memorys},'' {\em
  Proceedings of the IEEE}, vol.~106, no.~2, pp.~260--285, 2018.

\bibitem{dac2018-reagen}
B.~Reagen, U.~Gupta, L.~Pentecost, P.~Whatmough, S.~K. Lee, N.~Mulholland,
  D.~Brooks, and G.-Y. Wei, ``{Ares: A Framework for Quantifying the Resilience
  of Deep Neural Networks},'' in {\em Proceedings of the 55th Annual Design
  Automation Conference}, 2018.

\bibitem{frontiers2016-RPU}
T.~Gokmen and Y.~Vlasov, ``{Acceleration of Deep Neural Network Training with
  Resistive Cross-Point Devices: Design Considerations},'' {\em Frontiers in
  Neuroscience}, vol.~10, p.~333, 2016.

\bibitem{isca2016-chen}
Y.-H. Chen, J.~Emer, and V.~Sze, ``{Eyeriss: A Spatial Architecture for
  Energy-Efficient Dataflow for Convolutional Neural Networks},'' in {\em
  {International Symposium on Computer Architecture}}, 2016.

\bibitem{pieee2017-sze}
V.~Sze, Y.-H. Chen, T.-J. Yang, and J.~S. Emer, ``{Efficient Processing of Deep
  Neural Networks: A Tutorial and Survey},'' {\em Proceedings of the IEEE},
  vol.~105, pp.~2295--2329, Dec 2017.

\bibitem{zagoruyko2016-wideresnet}
S.~Zagoruyko and N.~Komodakis, ``{Wide Residual Networks},'' in {\em {British
  Machine Vision Conference}}, 2017.

\end{thebibliography}
}

\end{document}